\documentclass[preprint,12pt,authoryear]{elsarticle}

\usepackage{amssymb}
\usepackage{gensymb}
\usepackage{graphicx}
\usepackage{multirow}  
\usepackage{booktabs}       

\journal{Neural Networks}

\begin{document}

\begin{frontmatter}



\title{Learning Online Visual Invariances for Novel Objects via Supervised and Self-Supervised Training}


\author[inst1]{Valerio Biscione}

\affiliation[inst1]{organization={Department of Psychology},
            addressline={University of Bristol}, 
            city={Bristol},
            postcode={BS8 1TL}, 
            country={United Kingdom}}

\author[inst1]{Jeffrey S. Bowers}


\begin{abstract}
Humans can identify objects following various spatial transformations such as scale and viewpoint. This extends to novel objects, after a single presentation at a single pose, sometimes referred to as online invariance. CNNs have been proposed as a compelling model of human vision, but their ability to identify objects across transformations is typically tested on held-out samples of trained categories after extensive data augmentation. This paper assesses whether standard CNNs can support human-like online invariance by training models to recognize images of synthetic 3D objects that undergo several transformations: rotation, scaling, translation, brightness, contrast, and viewpoint. Through the analysis of models' internal representations, we show that standard supervised CNNs trained on transformed objects can acquire strong invariances on \textit{novel} classes even when trained with as few as 50 objects taken from 10 classes. This extended to a different dataset of photographs of real objects. We also show that these invariances can be acquired in a self-supervised way, through solving the same/different task. We suggest that this latter approach may be similar to how humans acquire invariances.
\end{abstract}





\end{frontmatter}
\tnotetext[1]{prova prova}

\section{Introduction} \label{Introduction}
Humans can identify objects despite the variable images they project on our retina, including variation in image size, orientation, illumination, and position \citep{Tanaka1996}. Critically, this invariance is computed `online': a transformed, unfamiliar object can often be recognized after a single presentation at a given pose. For example, after identifying a novel object that is projected at a given retinal location we can immediately identify the object across a wide range of retinal locations \citep{BlythingCommentary2021, Blything2021}. How the visual system succeeds under these conditions is still poorly understood, but we know that this task is solved through hierarchical processing along the ventral stream that ends at the inferotemporal cortex, where the activation of neural populations are largely independent to object transformation \citep{Bar1999}.

Recently, Convolutional Neural Networks (CNNs) have been proposed as a model for the human visual system \citep{YaminsDiCarlo2016, Khaligh-Razavi2014, SchrimpfBrainScore2018}. This is currently a topic of intense debate. On one hand, CNNs' internal activations are predictive of brain recordings in response to images taken from classic benchmark datasets in the mid and high levels of the ventral visual pathway \citep{Yamins2013, Yamins2014, Cichy2016ComparisonCorrespondence}. On the other hand, CNNs show some behavioural discrepancies that challenge their plausibility as a model of the visual system: they are susceptible to adversarial attacks \citep{Adversarial2013, Dujmovic2020}; they are highly susceptible to low amount of image degradation \citep{Geirhos2020}; they often classify images based on textures instead of shape \citep{Geirhos2020} and local instead of global features \citep{Baker2018, Malhotra2021}, and do not account for humans' similarity judgments of 3D shapes \citep{German2020CanJudgments}. Within the framework of this debate, we aim to answer the question of whether CNNs can learn to support online invariance for a wide variety of transformations commonly experienced in the human visual environment. This would help us better understand the limits and the possibilities of using CNNs as a model of the human visual system. 

\begin{figure}[!ht]
\centering
  \includegraphics[width=1\linewidth]{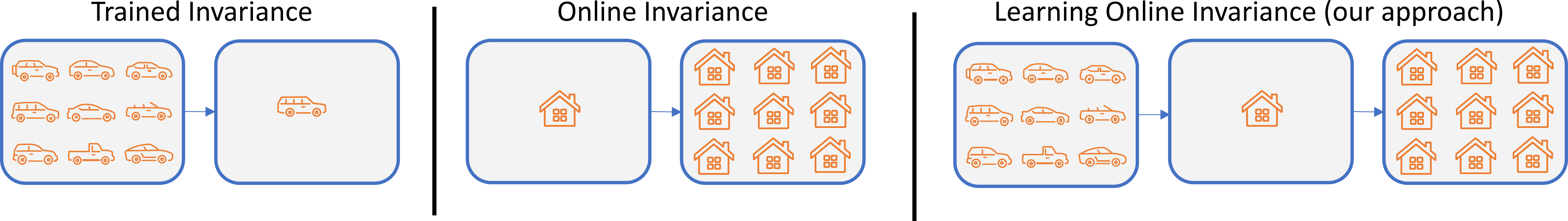}
\caption{Illustration of trained and online invariance and our proposal for learning online invariances. In this figure, the transformation we depict is translation, but the general logic applies to any transformations. \textbf{Left}: In trained invariance, after a model is trained on some classes of objects across different locations it can recognize novel instances from those same classes across different locations. This property is commonly obtained in CNNs through extensive data augmentation. \textbf{Middle}: In online invariance, a novel object can be instantly recognized across several transformations, after have experienced it at one location. Neural networks fail to show this property (see Section \ref{relatedwork}). \textbf{Right}: In our approach, a model is trained on one dataset of transformed objects in order to acquire online invariance to novel classes}
  \label{fig:FigExpl}
   \end{figure}

When comparing CNNs to human vision, it is important to  distinguish between \emph{online} transformation invariance and \emph{trained} transformation invariance (Figure \ref{fig:FigExpl}). Trained invariance refers  to  the  ability to  classify  novel exemplars of objects from \emph{trained} classes in  trained  locations  (e.g., invariance to the pose of a specific image of a dog after exposure to multiple poses of other exemplars of dogs). This is the standard approach in training CNNs, and because training is rarely performed through the lens of human cognition, it often includes augmentations that are unlikely to be experienced by humans (e.g. color jittering) as well as exclude common ones (e.g. change in viewpoint). 
Accordingly, the trained invariance commonly reported in CNNs is qualitatively different from human \emph{online} invariance that extends to \emph{novel objects from unseen classes} following exposure to only one or a few poses \citep{Bowers2016}. 
The degree to which CNNs are invariant to online transformation is unclear. 

In this work we aim to close this gap by testing, with a variety of methods, whether CNNs support online invariance for novel objects from novel classes even when the novel classes come from a new dataset. We only trained and tested models using naturalistic transformations, namely, scale, rotation, translation, brightness, contrast and change in viewpoint. Particular attention is given to the latter as it is not a simple transformation of a 2D image and thus is not typically used when performing data augmentation. We show that common CNNs models can acquire, to a strong degree, all the above invariances, even when trained with as few as 50 images of objects taken from 10 categories. Critically, these models not only support invariance for trained object classes, but also for untrained object classes. We find this following supervised learning, and also in a self-supervised manner by classifying pairs of objects as the same or different. We argue that this latter approach is more psychologically plausible, and discuss the implication in terms of human psychology in Section \ref{signPSY}. 

\subsection{Related work} \label{relatedwork}
 
The distinction between trained and online invariance has been the focus of recent work on translation invariance (recognizing an object at novel retinal locations after having seen it at one location). Behavioral studies have highlighted the extent to which humans possess online translation invariance \citep{BlythingCommentary2021, Blything2021}, and it is often claimed that convolutional and pooling operations endow CNNs a similar degree of online invariance to translation \citep{DeepLearningMarcus2018, Lecun1995}. However, several experiments have shown that this is not the case, and indeed, CNNs not only fail to support online invariance to translation, but also fail to support online invariance to scale, rotation, and flipping along the vertical and horizontal axis \citep{Kauderer-Abrams2017a, Gong2014, Blything2021, Chen2017}.
One approach to achieving online invariance in CNNs is through architectural modification, for example, adding a Global Average Pooling layer to the end of the convolutional block results in complete translation invariance \citep{Blything2021}. Additional architectural modifications have been introduced in order to support other types of transformation. This has resulted in a proliferation of modified models that accounted for either individual transformations (e.g. \citealt{Xu2014scale, Han2020b} for scale; \citealt{Kim2020CyCNN:Layers, Marcos2016} for rotation) or multiple transformations at the same time \citep{cohen2016}. 
Although these models may prove valuable in terms of technological advancements in the field of object recognition, the architectural modifications are not guided by any concern of biologically plausibility.  And more importantly, a key assumption of these approaches is that standard CNNs architectures are incapable of supporting online invariances. If it turns out that some training regimes allow standard CNNs to support online invariances, the introduction of additional architectural (innate) mechanisms may not be required either for engineering or psychological considerations.  

\cite{Biscione2020} and \cite{Blything2021} first reported examples of a classic CNN model (VGG16), without any architectural changes, exhibiting complete online invariance to translation when the model was pretrained on the ImageNet dataset. That is, a network pretrained on ImageNet supported translation invariance to other, very different datasets (e.g. MNIST, EMNIST, and others). A key feature of the pretraining was that it consisted of resizing and crop augmentations, which indirectly resulted in translated versions of the same images. That is, these findings lend support the hypothesis online invariance can be trained by using a dataset of translated samples. \cite{Biscione2021JMLR} further investigated this approach by pretraining several convolutional architectures on a variety of datasets that differing in the complexity of the translated stimuli. They observed that that training on simple translated datasets was often enough to support online translation invariance on other, \emph{novel} classes, sometimes very different from the trained ones.

Overall, a review of the literature points to the following facts: CNNs are not, by design, architecturally invariant to any tested transformation (scale, rotation, translation). Many \emph{ad hoc} architectural changes can be employed to provide invariance for some transformations. Only recently it has been found that pretraining standard CNNs (without any architectural modification) with translated objects endows them with online translation invariance (that is, translation invariance for classes \emph{different} than the pretrained ones). However, it has not yet been explored to what extend this technique can be employed across transformations other than translations and across different networks architectures. The main goal of the present work is to fill this gap by testing a large set of transformations (in fact, a set covering all possible transformations in 3D space) on many networks, both supervised and a self-supervised.

\subsection{Outline of the Current Work}
In this work, we extend the work of \cite{Blything2021}, \cite{Biscione2020}, and \cite{Biscione2021JMLR}, and show that: 1) online invariance to many different types of transformations can be acquired with no architectural change by pretraining the network on the same transformations assessed at test; 2) many different supervised networks acquire this property, as does a self-supervised network that solves the same/different task.  We argue that the latter task provides a more psychologically plausible approach to learning invariances in infants. 

We tested 7 supervised networks (AlexNet, VGG11, VGG19, Resnet-18, Resnet-50, GoogLeNet, Densenet-201) and one self-supervised network (detailed in Section \ref{samediffnet}) on 6 invariances: translation, rotation, scale, brightness, contrast, and viewpoint. Although this final transformation does not belong to standard data augmentation technique (in that it is not a transformation of a 2D image), it is a psychologically relevant transformation, normally experienced by an observer in a 3D world. In order to obtain different viewpoints samples, we generated a new dataset based on ShapeNet (Section \ref{multiviewds}).

Our general approach consisted in training each network on 10 classes from the ShapeNet dataset, for each transformation separately, and without any transformation.  We then tested on 10 novel classes either from ShapeNet or from a dataset of photographed objects, ETH-80. A network trained with a certain transformation was tested on the corresponding transformation (e.g. a network trained with translated ShapeNet objects was tested on translated objects from novel classes). The networks trained without any transformation were tested across all transformations.
With the self-supervised network, network performance on novel classes can be tested directly without having to retrain the model (Section \ref{invsamediff}). For the supervised networks, assessing the amount of online invariance on novel classes through the classification performance requires retraining on the novel classes. This in turn could result in catastrophically forgetting the acquired invariances (\citealt{Biscione2021JMLR} as discussed in more detail in Section \ref{invallmod}). Therefore we assessed the amount of online invariance with two approaches that did not require retraining: a 5-alternative forced choice task, in which the internal representation was used to identify which transformed object, among five candidates, was the same as the target object (Section \ref{5AFC}); a direct measure of the internal representation similarity between transformed and un-transformed samples (Section \ref{repran}). These tests were always performed on novel classes. In Section \ref{CrossTr}, we also ran a cross-transformation analysis, in which we tested whether pretraining on one transformation afforded invariance to any other transformation. In Section \ref{MultiObj} we investigated how the number of trained objects impacted the acquisition of invariances to transformations on novel classes of objects.  A summary of the result is provided in Section \ref{summary}. Finally, the limitation of the current method and the implications of our findings for Machine Learning and Psychology are discussed in Section \ref{Discussion}.


\section{Methodological Overview} \label{Methods}


\subsection{Transformation Scheme}
We explore whether each model can learn each of the 6 invariances by applying a corresponding stochastic transformation scheme $T_{t,\theta}$ to each object, where $t$ identifies a transformation with parameters $\theta$ (e.g. $t= rotation$ and $\theta = [-180, 180]$ indicates a random rotation on the image plane between -180° and 180°. 
When $t \neq viewpoint$, the object is presented at an inclination of 80\degree{} and azimuth of 36\degree{}. The parameters $\theta$ for each transformation $t = \{rotation, scale, translation, brightness, contrast, viewpoint\}$ are shown in Figure \ref{fig:FigSphere}B.

\subsection{Multi-viewpoint Dataset} \label{multiviewds}
\begin{figure}[!ht]
\centering
  \includegraphics[width=0.9\linewidth]{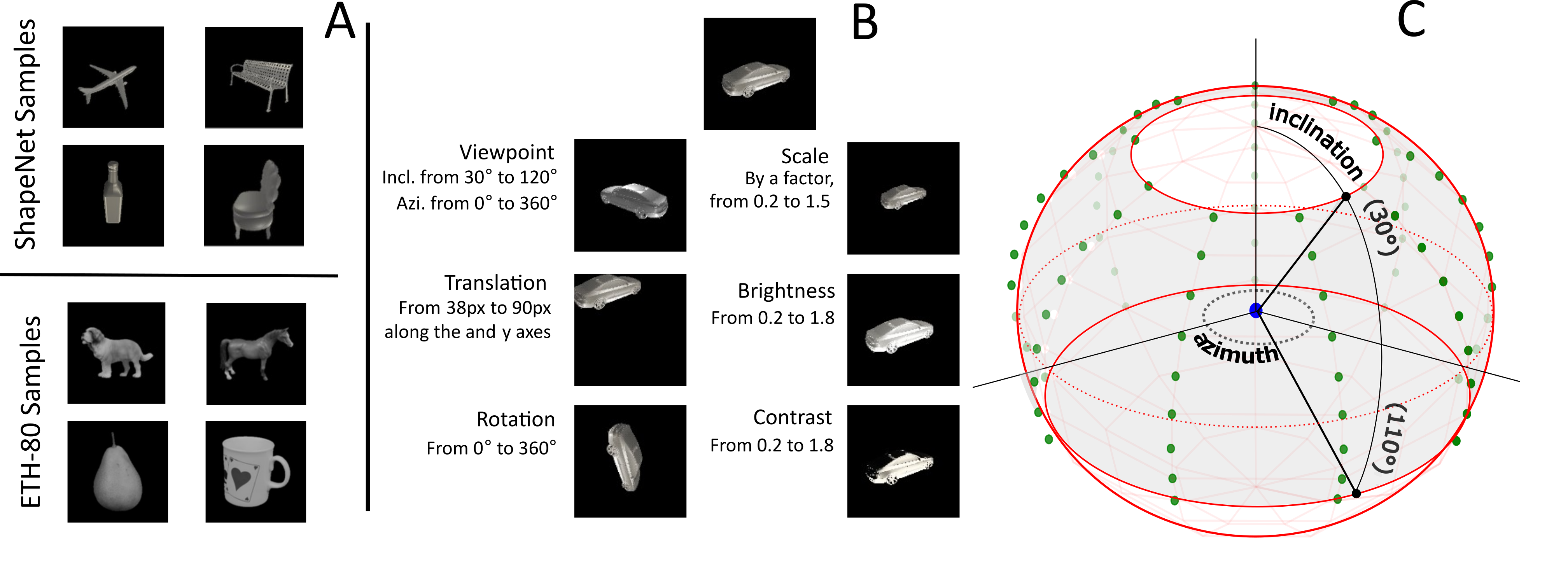}
\caption{\textbf{A}: Some samples from the ShapeNet and ETH-80 datasets (from one viewpoint and without any 2D transformations). \textbf{B}: Example of different samples resulting from stochastically applying a transformation to the 3D object, and the parameters used for each transformation. \textbf{C}: Representation of the method used for generating different objects' viewpoints. The cameras (green dots) were placed at different locations around a sphere. Each camera pointed at the center of the sphere, where the 3D object was placed (blue dot). The cameras were only placed between an inclination of 30\degree{} to 110\degree{}, and covered the whole sphere longitudinally (shaded area).}
  \label{fig:FigSphere}
  \end{figure}

In order to investigate whether invariance to viewpoint can be acquired we generated a novel 2D dataset based on ShapeNet \citep{Chang2015shapenet}, a dataset containing 50300 3D objects across 55 categories. Each 3D object was rendered textureless on a uniform black background as a 128x128x3 image. To obtain variation in viewpoint, we placed a  camera on a sphere at an inclination ranging from 30\degree{}   to 110\degree{} (10\degree{} intervals) and an azimuth covering the whole sphere (36\degree{} intervals, Figure \ref{fig:FigSphere}C), generating 90 different viewpoints per object. In the following sections the term``object" refers to any image of a specific 3D model regardless of viewpoint or 2D transformations applied to it. Notice that this approach differ greatly from other approaches in which ShapeNet's 3D objects are used in networks that directly consume point clouds (e.g. PointNet, \citealt{PointNet}, 2017).

\subsection{Same/Different Task Approach} \label{samediffnet}
Consider a set of objects $O=\{o_n\}_{n=1}^N$ subjected to a transformation scheme $T_{t,\theta}(o_n)$ which stochastically produces an image $x_n^i$. We define $X_n=\{x_n^i\}_{i=1}^M$ as a set of images resulting from applying $M$ times the transformation scheme $T_{t,\theta}$ to the object $o_n$. 
We furthermore define an embedding module $g_\psi(\cdot)$ which produces a compact representation $z_n^i$ of an input  image $x_n^i$. Pairs of inputs $\{x_n^i, x_m^j\}$ are build such that either they are from the same set $X_n$ and thus $n = m$ (they represent same object $o_n$) with probability $p$ or they are from two different sets so that $n \neq m$ (they represent different objects) with probability $1-p$. The pair embeddings $\{z_n^i, z_m^j\}$ are aggregated through $a(\cdot, \cdot)$ and fed into $h_\phi(\cdot)$, as a non-linear function approximator parameterized by a learneable weights $\phi$ which returns a dissimilarity score $\hat{y}$. We use mean square errror (MSE) for the loss, matching the dissimilarity score with object identity \citep{Sung2018a}:

$$\mathcal{L}(\hat{y}_{n,m}) = \sum^{N}(\hat{y}_{n,m} - \mathbf{1}\{n \neq m\})^2$$

This is minimum when the $\hat{y} = 1$ for different objects and $\hat{y} = 0$ for the same objects. 

\begin{figure}[!ht]
\centering
  \includegraphics[width=1\linewidth]{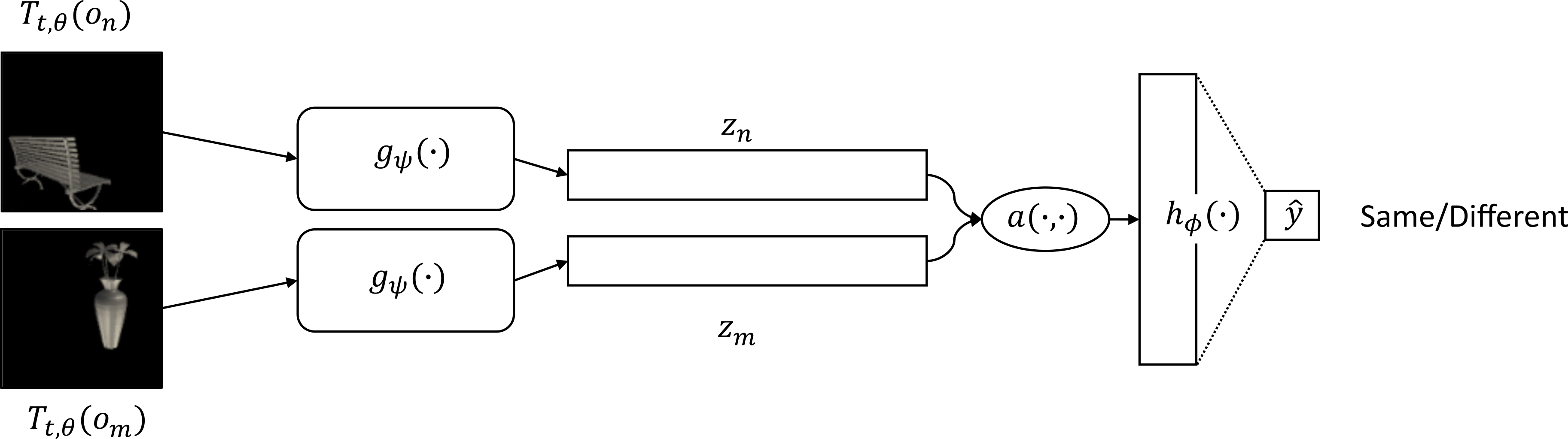}
\caption{Overview of the proposed Same/Different Network. See text for details.}
  \label{fig:FigModel}
   \end{figure}
   
In particular, the embedding module $g_\theta$ is a a copy of VGG11 in which the last fully connected layer (the classifier) is replaced by another fully connected layer which output is an embedding of 512 units $+$ Sigmoid activation function. As an aggregation function we use $a(z_n^i, z_m^j)=|z_n^i - z_m^j|$. The dissimilarity module $h_\phi$ is a single fully connected layer with a single output unit $+$ Sigmoid. The probability $p$ of a pair being composed by the same objects was set at $0.5$.

Our approach is related to self-supervised learning, in which a system is trained with unlabeled data on a surrogate task to acquire features that can be used on a downstream task \citep{JingTian2019}. For example, SimCLR, \citep{chen2020simple}, and Relational Reasoning Network \citep{Patacchiola2020} use mini-batches of augmented samples and map together representations of different augmentations of the same image and separate the representations different samples. One of the main differences between our self-supervised approach on one hand, and SimCLR and Relational Reasoning Network on the other, is that we use a 3D model instead of a 2D image to generate data augmentations, and thus we can consider augmentations that are normally excluded from other models, that is viewpoint. This could be expanded to changes in lightning, textures, and other material/environmental properties of the 3D object, allowing us to train models on much more naturalistic data. However, our aim is not to compare our implementation with similar contrastive approaches or to obtain a state-of-the-art accuracy, but to show as a proof of concept that a simple self-supervised network can indeed acquire these invariances.

\subsection{Supervised Approach}
Supervised training was arranged as usual: considered a set of categories $C=\{c_k\}_{k=1}^K$  each containing a number of objects $O_k=\{o_{n,k}\}_{n=1}^N$ subjected to a transformation scheme $T_{t,\theta}(o_{n,k})$ which stochastically produces an image $x_{k,n}^i$. In a supervised approach, the identity of an object $n$ is discarded and only the class $k$ is considered during training. A non-linear function approximator $f_\phi(x_k^i)$ parameterized by lerneable weights $\phi$ is trained to minimize the error in predicting the output class $k$. 
For $f_\phi$ we consider a variety of Convolutional Neural Network: Alexnet, VGG11, VGG19, Resnet-18, Resnet-50, GoogLeNet, Densenet-201, covering a wide range of feedforward architectures. 
Each network was trained through stochastic gradient descent on the cross-entropy loss.

\section{Experiments and Results} \label{Exp}
We assessed trained and online invariance in six different supervised networks and in the self-supervised same/different network.  We trained all networks on the same samples obtained from the ShapeNet objects dataset: we randomly sampled 250 3D objects  from 10 classes, totalling $2500$ objects. The classes were randomly selected to be: bench, bottle, bus, clock, faucet, jar, knife, laptop, rifle, table. Each model was trained on two conditions: (1) with each transformation (e.g., translation, rotation, etc.) applied separately, and (2) without transformations. This results in models in this latter condition being trained on less samples (as the objects used in this condition were not transformed) but this does not affect the results, as the analysis in Section \ref{MultiObj} shows. We used Adam optimizer with learning rate $10^{-3}$, batch size of 64. We trained until convergence, stopping when the exponential moving average of the loss  (with $\alpha=0.1$) computed on the training set did not decrease by at least $0.01$ for 250 mini-batch iterations. We normalized the input images between $-1$ and $1$ by pre-computing the mean and standard deviation for a large subset of the ShapeNet/ETH-80 datasets.    
Each experiment is repeated over 3 random seeds.

To measure the degree of online invariance we selected 3D models from 10 novel classes from ShapeNet: airplane, bathtub, car, mobile-phone, chair, guitar, lamp, pot, sofa, vessel.  We also tested online invariance on a new dataset, ETH-80, which includes photographs of 80 real life objects along 8 classes (apple, car, cow, cup, dog, horse, pear, tomato), each object photographed from 41 viewpoints, which made it ideal for our viewpoints tests (Figure \ref{fig:FigSphere}). The background was removed from these images, and they were converted to grayscale values. We excluded the class ``tomato'', as it was almost indistinguishable from the class ``apple'' when converted to grayscale.

\subsection{Invariance in the Same/Different Network} \label{invsamediff}
In the case of the Same/Different Network, invariance is manifest when the model outputs a low dissimilarity score to transformations of the same object and a high score to different objects, and online invariance can be directly measured by presenting novel classes. Each version of the same/different network was tested on the same transformation it was trained on (apart for the networks trained without any transformations, which were tested on each transformation separately).  With both novel ShapeNet classes and ETH objects, the performances of the networks trained on transformations was consistently better than the network trained on untransformed objects, with performance $>$ 90\% in all conditions other than viewpoint (Table \ref{samediff-t}). Testing on the ETH dataset resulted in a drop in accuracy of around 4\% for all but the viewpoint transformation, which dropped by $\sim16\%$. This is still noteworthy considering the difference in appearance between the trained ShapeNet objects and these ETH objects and the low variability within ETH classes (e.g. when a pair is made of two different objects from the same class, such as two mugs, the low inter-class variability makes it more difficult to classify the objects as different). Overall, the network show strong online invariance for novel classes and novel datasets.

\begin{table}[]
\caption{Accuracy for the Same/Different Network on novel ShapeNet classes and ETH-80}
  \label{samediff-t}
  \centering
\resizebox{\textwidth}{!}{%
\begin{tabular}{llllllll}                                                                                             &                                                                            &            &             & \multicolumn{3}{l}{\textbf{Transformations on Test}} &            \\ \cmidrule(r){3-8}                                                                     &                                                                            & Viewpoint  & Translation & Rotation         & Scale           & Brightness      & Contrast   \\
\multicolumn{1}{c}{\multirow{2}{*}{ShapeNet$\rightarrow$ShapeNet}} & \begin{tabular}[c]{@{}l@{}}Trained with \\ Transformations\end{tabular}    & 82.65$\pm$0.75 & 95.02$\pm$0.88  & 96.61$\pm$0.49       & 95.06$\pm$0.35      & 98.54$\pm$0.36      & 97.85$\pm$0.39 \\
\cmidrule(r){2-8}
\multicolumn{1}{c}{}                                                 & \begin{tabular}[c]{@{}l@{}}Trained without\\  Transformations\end{tabular} & 50.85$\pm$0.59 & 50.56$\pm$1.35  & 50.58$\pm$1.41       & 50.68$\pm$1.37      & 50.01$\pm$1.19      & 52.17$\pm$1.19 \\ \midrule

\multirow{2}{*}{ShapeNet$\rightarrow$ETH-80}                                        & \begin{tabular}[c]{@{}l@{}}Trained with \\ Transformations\end{tabular}    & 66.67$\pm$3.56 & 93.33$\pm$1.78  & 92.85$\pm$3.08    & 92.85$\pm$4.04      & 94.76$\pm$3.56      & 93.33$\pm$3.56 \\                                        \cmidrule(r){2-8}                             & \begin{tabular}[c]{@{}l@{}}Trained without\\ Transformations\end{tabular}  & 53.81$\pm$2.43  & 51.42$\pm$5.83  & 51.90$\pm$5.87 & 53.33$\pm$5.51       & 54.76$\pm$4.71      & 54.28$\pm$4.04

\end{tabular}}
\end{table}

\subsection{Invariance Across All Models} \label{invallmod}

Assessing online invariance on supervised networks presents special challenges. One approach consists of retraining the networks on un-transformed samples from novel classes, and then measuring the classification performance on the transformed version of the same classes. For example, to assess translation invariance, a network could be trained on samples from novel classes, always presented at the center of the canvas, and then tested on the same classes presented at different locations. The resulting classification accuracy would correspond to the degree of online invariance to translation. The problem with this approach is that it requires retraining on novel classes, which might result in the catastrophic forgetting of the acquired invariances \citep{French1999CatastrophicNetworks}. This means that even if the networks did acquire the invariance from the transformed dataset, we would not be able to observe it through the classification performance alone \citep{Biscione2021JMLR}. This point will be further discussed in Section \ref{signML}.

Following \cite{Biscione2021JMLR} we opted to assess online invariance by measuring the similarity of internal representation between transformed and untransformed versions of the same objects in both the same/different and supervised networks using cosine similarity:

$$C^l(a, b) = \frac{d^l(a) \cdot d^l(b)}{\left \| d^l(a) \right \| \left \| d^l(b) \right \|},$$

where $d^l(a)$ is the activation at layer $l$ given input $a$.
We then assessed invariance in two ways, namely, in a choice task in which similarity scores were used to classify objects, and a normalized cosine similarity score that accounted for within vs. between objects cosine similarity scores. We consider these two measures in turn. 

\subsubsection{5-Alternative Forced Choice Task} \label{5AFC} 
\begin{figure}[!ht]
\centering
  \includegraphics[width=1\linewidth]{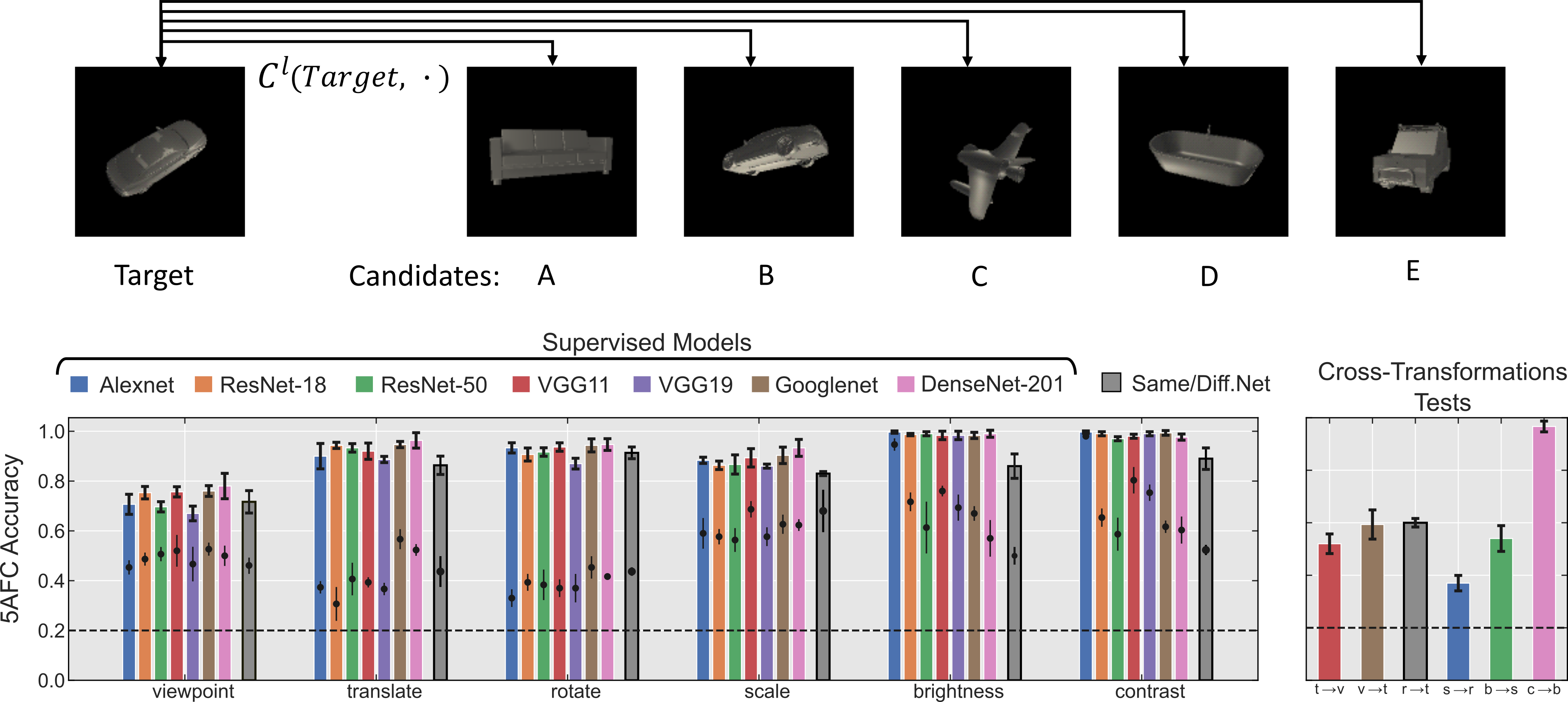}
\caption{\textbf{Top}: Illustration of the 5-alternative forced choice task. One object is compared with 5 alternatives in which one of them is the same object (in the picture, Candidate B). All samples are subjected to a transformation (that varied by condition). We used the cosine similarity of the internal representations across pairs of Target and Candidates as a metric of choice. \textbf{Bottom-left}: Results in terms of accuracy for each transformation and each model. The bar represents the accuracy for a network trained on the same transformation it is been tested on. Compare this with the black circle inside each bar which represents the accuracy for a network trained on un-transformed objects.  \textbf{Bottom-right}: accuracy when a network trained on one transformation is tested on a different transformation. Apart from brightness and contrast, training on one transformation did not generalize to other transformations. Overall, these plot show a high degree of learnt online invariance to a transformation when the network is trained on the same transformation (the classes used for the 5AFC were not used during training)}
  \label{fig:FigBarplot}
  \end{figure}

We tested each network on 5-alternative choice task (5AFC). The cosine similarity between a target object and 5 candidates was computed, setting one candidate to be the same as the target (Figure \ref{fig:FigBarplot}, top). All objects were subjected to random transformation, and each network was tested on the same transformation it was trained on. Objects were randomly sampled from the unseen classes. We ran 100 trials for each condition, and defined as ``correct" a response trial in which the cosine similarity at the last layer was higher for the transformed version of the target object. For the Same/Different task, the cosine similarity was computed on the embedding vector. The results are shown in Figure \ref{fig:FigBarplot}, bottom-left,  together with the results of the same test applied on a network trained on un-transformed samples (black circles in the plot). The results indicate high performance in all transformation conditions, with viewpoint being the most challenging. That is, all models acquired an impressive degree of online invariance. The 5AFC test repeated on the ETH objects resulted in a slight drop in accuracy (by $\sim16\%$), with the viewpoint transformation suffered the most extreme drop in accuracy (lowered to $\sim60\%$, and thus still remarkably higher than chance) compared to the ShapeNet objects (see Supplementary Materials). However, even with this new dataset, accuracy was always significantly higher with network trained on transformed samples than pretrained on un-transformed samples.

\subsubsection{Cross-Transformation Tests} \label{CrossTr}
One possible interpretation of these findings is that the online invariance for a given transformation type emerges in response to training on the corresponding transformation, and another is that training on any invariance will improve all forms of online invariance. To test this, we performed cross-transformations tests: a network trained with a transformation $t$ was subjected to a choice-test with a different transformation applied. For the viewpoint, translation, rotation, and scale, the obtained performance is similar to the performance of an un-transformed network (Figure \ref{fig:FigBarplot}, bottom-right panel). On the other hand, brightness and contrast generalized to each other, and only to each other: for example, training translation did not increase invariance on either brightness or contrast. 

\subsubsection{Internal Representations Analysis} \label{repran} As a further metric of transformation invariance, we directly explored how the internal representations of an object changed when transformed compared to a base view. Consider $\tau_t^\theta(o_n)$ as a specific transformation $t$ with parameter $\theta$ of object $o_n$, (e.g. rotation of the object $o_n$ by $60\degree$). Let us define a set of baseline parameters $\alpha$ for each transformation $t$:  $1$ for scale, brightness, and contrast; 0\degree{} for rotation, $(64, 64)$ pixels for translation (center of the canvas), inclination$=$80\degree{} and azimuth$=$ 36\degree{} for the viewpoint transformation. Therefore, for each network trained on a transformation $T_{t,\theta}$, and a set of randomly sampled objects $\{o_1, ..., o_R\}$ from novel classes, we defined $I_t(\theta)$ (for \emph{Invariance}) as the average similarity of representations between the objects base-view and their transformations $\tau_t^\theta$. A network could obtain a high Invariance score by simply collapsing all representations together, regardless of variability across objects or class identity. The model could be trivially invariant to object transformation by being  invariant to any object feature. We accounted for this possibility by measuring the variability across \emph{different} objects: we define $U_t(\theta)$ (for \emph{Uniformity}) as the between-objects invariance, that is the total average similarity across different randomly sampled objects
$u = \{u_1, ..., u_n\}$ and $v = \{v_1, ... , v_n\}$:

$$I_t(\theta) = \frac{1}{R}\sum_r^R C^l(\tau_t^\alpha(o_r), \tau_t^\theta(o_r)) \quad\mathrm{and}\quad U_t(\theta) = \frac{1}{N}\sum_n^N C^l(\tau_t^\alpha(u_n), \tau_t^{\theta}(v_n))$$

We used $U_t(\theta)$ as a baseline invariance across objects: if a model has learnt to be non-trivially invariant to transformation $t$, $I_t(\theta)$ should be higher than $U_t(\theta)$. Therefore we defined the Adjusted Invariance Metric as

$$ \tilde{I}_t(\theta) = \frac{I_t(\theta) - U_t(\theta)}{1-U_t(\theta)},$$

which returns the adjusted invariance for each transformation by parameter $\theta$. A value of $1$ indicates identical representation between transformed version of the same objects (same internal representation across transformed versions of the same objects, and different internal representations of different objects); a value of $0$ indicates that two different transformations are as different as two different objects.  

\begin{figure}[ !ht]
\centering
  \includegraphics[width=1\linewidth]{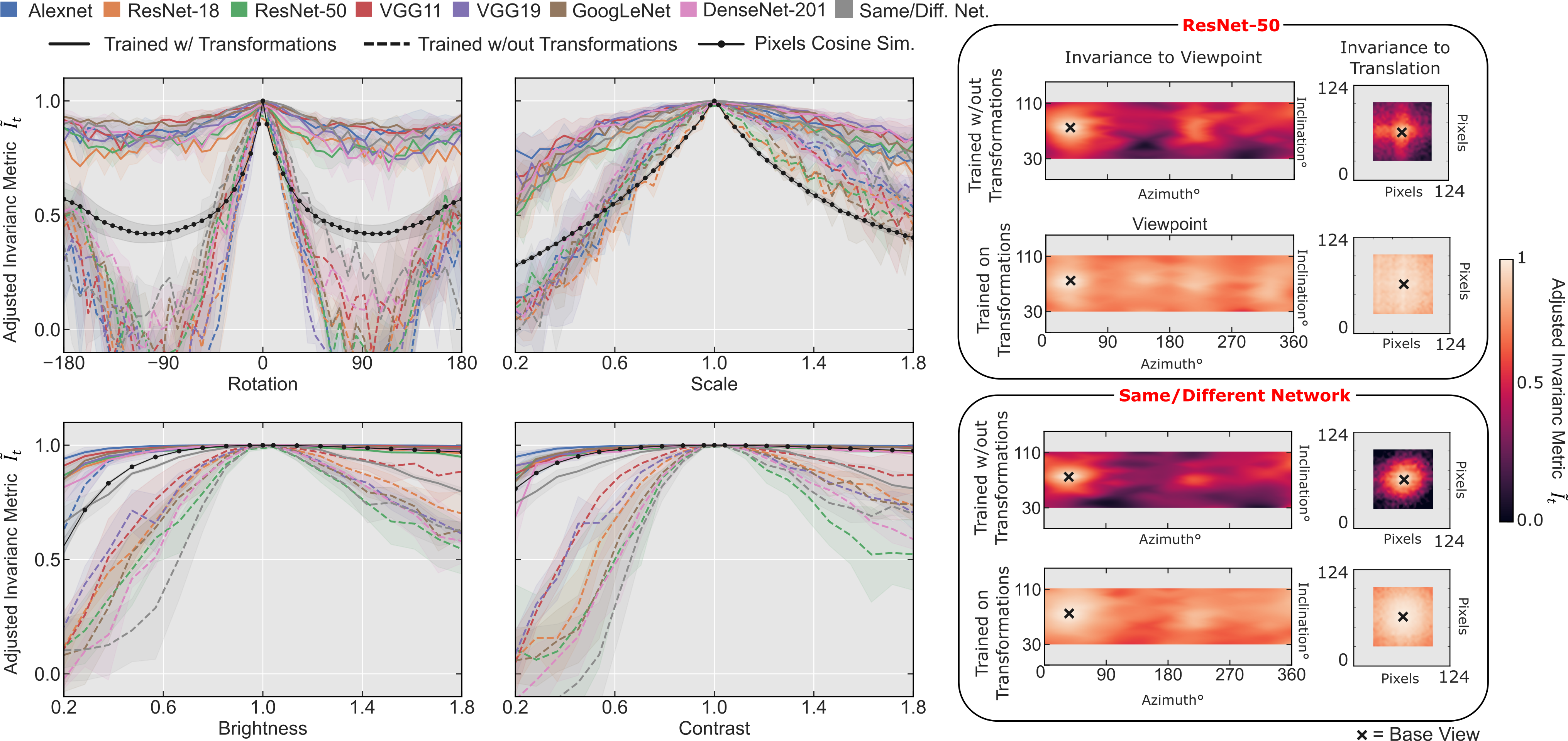}
\caption{Invariance Representation Metric between a base view of an object and its transformed versions.  The networks trained with transformations (continuous lines) have a high invariance across most levels of each transformation, whereas for networks trained on un-transformed objects (dashed lines) the similarity sharply decreases even at low levels of transformations. The results in the four panels on the \textbf{left} are averaged across $200$ random samples from untrained classes. The shaded area represent the standard deviation across the 3 seeds for each network. The black line with circular markers represents the cosine similarity at the pixel level at different transformations. The \textbf{right} panels show the results of viewpoint and translation invariance for two networks (other networks are shown in \ref{SUPPinvanalysis}). The results were consistent across all tested networks. }
  \label{fig:FigInvariance}
   \end{figure}

All objects used in this analysis belong to novel, untrained classes from ShapeNet. The results (Figure \ref{fig:FigInvariance}) clearly show that, when trained and tested on corresponding transformations, each model can acquire a strong invariance to that transformation, even for novel objects and classes. This is not the case with networks not trained on transformed samples, as their representations strongly depend on the object pose. There does not seem to be any relevant difference amongst models, either supervised or self-supervised. The analysis of $U_t(\theta)$ (in Supplementary Materials) also shows that the networks not trained on transformed samples collapses representations of different, transformed objects together. For these models, a car rotated 90\degree{} is very similar to a horse rotated 5\degree{} (however, un-transformed objects were still clearly separated from different, un-transformed objects).  

\subsection{Influence of Number of Objects Trained} \label{MultiObj}
\begin{figure}[ !ht]
\centering
  \includegraphics[width=1\linewidth]{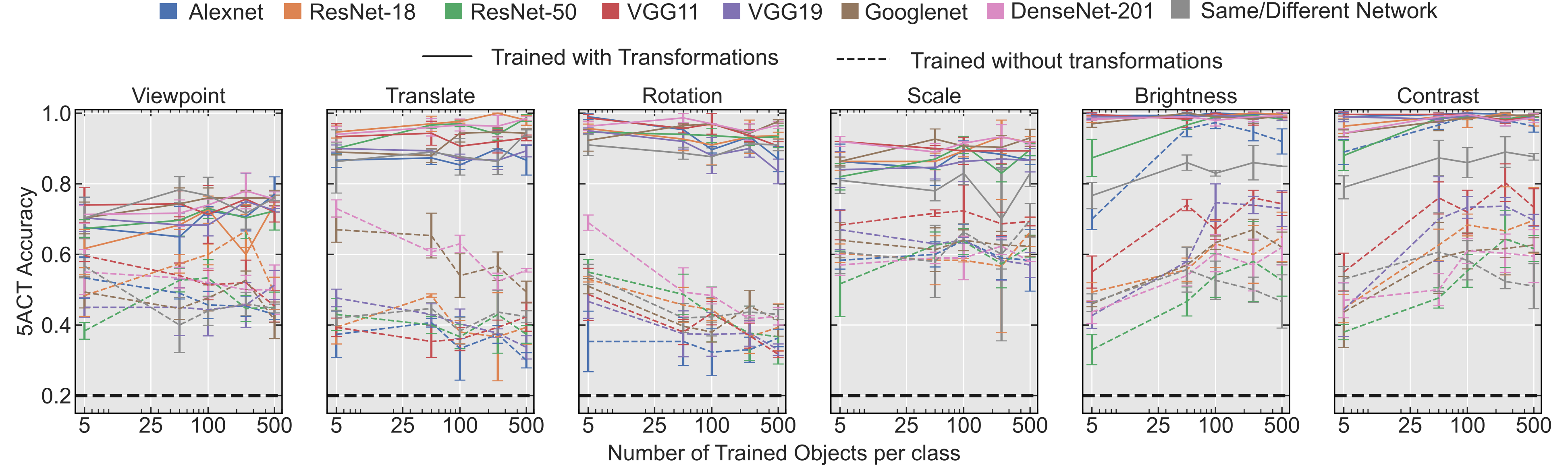}
\caption{Effect of training on different number of objects on the learnt invariance. There is no effect on the network trained on transformed objects. However, there seems to be some effect on network trained on un-transformed objects, but the direction of this effect is not consistent across transformations.}
  \label{fig:FigMultiobj}
   \end{figure}

In the preceding experiments, each network reached a high level of invariance following training on $250$ objects per class. We explored how invariance changed as a function of the number of trained objects. For each network, we repeated the previous training with $N = \{5, 50, 100, 500\}$ where N is the number of objects per each category, and computed the results on the 5AFC task.
Surprisingly, there does not seem to be any effect on the number of trained objects with the transformed networks (Figure \ref{fig:FigMultiobj}): as few as 5 objects per classes (with a total of 50 transformed objects trained on) were enough to acquire strong invariances. The plot also shows that the invariance acquired was not due to the higher number of samples experienced by the network trained with transformations: increasing the number of objects (and thus the number of different samples) did not improve online invariance on networks trained on un-transformed samples. In fact, it appeared to decrease it in some cases. 

\subsection{Summary of the Results} \label{summary}
CNNs architectures can learn to support online invariance for scale, rotation, translation, brightness, contrast, and, to a lower amount, viewpoint, when trained on an appropriately transformed dataset. The findings are robust across a variety of architectures, learning setups (supervised and self-supervised), and training size (trained on as few as 50 objects taken from 10 classes). 

In the 5 alternative forces choice task (Section \ref{5AFC}) all networks performed at $\sim90\%$ accuracy for all transformations but viewpoint ( $\sim70\%$ across all networks), when tested on novel classes from ShapeNet. When tested on a novel dataset (ETH-80), accuracy dropped only slightly for all conditions but viewpoint invariance, which dropped to $\sim 60\%$, still remarkably higher than chance. Overall, viewpoint invariance appeared to be the most difficult transformation to generalize across novel classes. Similar results are obtained by analysing the networks' internal representations (Section \ref{repran}). We also verified that, in order to acquire invariance to a certain transformation, the networks needed to be trained specifically on that transformation, and that training on other transformations will not suffice (with the exception of contrast and brightness, which would generalize to each other, Section \ref{CrossTr}). Similar results are obtained when varying the number of trained objects, and networks were able to acquire all invariances following training on as few as 50 objects taken from 10 classes (Section \ref{MultiObj}). 

Overall, we found that training a network on a transformation on one set of objects allowed CNNs to support the corresponding transformation for novel objects taken from novel classes. For example, we showed that training rotation invariance on images of synthetic 3D buses supports rotation invariance on images of synthetic 3D chairs. We show that this extends to novel classes from the same synthetic dataset and on a different dataset of real, photographed objects (ETH-80).

\section{Discussion} \label{Discussion}
The finding that a range of visual invariances can be acquired using a general learning mechanism is significant both for the machine learning community and for the psychological literature.  We discuss both points in turn.

 \subsection{Significance for Machine Learning} \label{signML}
 As outlined in Section \ref{relatedwork}, the common approach to achieve strong invariance to various transformations is to apply architectural modifications to standard CNNs. Some of these modifications afford invariance to only one transformation (e.g. \citealt{Xu2014scale, Han2020b} for scale; \citealt{Kim2020CyCNN:Layers, Marcos2016} for rotation), others to multiple transformations \citep{cohen2016}. Our work shows that these modifications may not be necessary to acquire invariances. Indeed, we found through measuring internal representation that standard CNNs show strong invariances to a range of transformations following the appropriate training. Furthermore, the current approach allows the possibility of acquiring invariances that are not limited to 2D affine transformation (as in the architectural approach) but can be extended to all types of variations: we have shown here viewpoint variation, but it is possible to design a dataset to induce invariance to background, clutter, texture, etc. It is not clear how to modify a network's architecture to innately capture these invariances.

However, it is important to emphasize that these learnt latent invariances will not always manifest themselves in performance. For example, imagine that we want to perform classification on some classes, e.g. different breeds of dogs, and for computational reasons we only want to train with the dogs at one location and wish the network to be able to classify the dog at any location (that is, we want the network to possess online invariance to translation). With our approach, we would need to pretrain the network on a translated dataset, and then retrain the network on our dataset of dogs. The problem of this approach is that the re-training session could result in catastrophic forgetting \citep{McCloskey1989, French1999CatastrophicNetworks}, a phenomenon in which acquired capability rapidly degrade as the system learns a new task. In our cases, it means that even though CNNs can acquire online invariances, and these can be generalized on object from \emph{novel} classes (as we have shown through internal representation analysis), \emph{retraining} to perform classification could result in losing the learnt invariance.

Still, there are reasons to think this limitation can be overcome.  \cite{Biscione2021JMLR}, in the context of translation invariance, found that learnt invariances have higher chance to be retained if the re-training dataset is less complex than the pretraining one. Furthermore, there is ongoing work in overcoming catastrophic interference through interleaved training \citep{Schaul2015PrioritizedReplay}, selective plasticity \citep{Beaulieu2020, RusuProgressive, Fernando2017PathNet:Networks}, and by incentivizing sparse or disjoint representations \citep{French1999CatastrophicNetworks, LiuRotate}. Even though the current applicability of this approach \emph{as-is} in classification is limited, we suggest that promising advances in the field of catastrophic could provide a possible path for making use of the learnt invariances and thus complement or replace the current architectural approach. Our work highlights how the solution to catastrophic interference may have implications for how CNNs can solve various online invariances.

 We also note that the current results appear to be inconsistent with two recent findings. \cite{Hernandez-Garcia2019LearningCategorization} found low similarities of the latent embeddings of transformed objects for a network trained on augmented CIFAR10 (with random affine transformations, brightness and contrast). We believe that this is due to their reliance of Euclidean distance which underestimated the degree of scale invariance and subsequently (since the transformations were aggregated) the strength of all invariances. Also \cite{Xu2021} found inconsistent embeddings across classes of translated objects, which we believe to be due to the different training setup they employed. We expand on this in the Supplementary Material.  
 

\subsection{Significance for Human Psychology}   \label{signPSY} 
An important observation regarding invariances in humans is that they exist from an early age:  rotation \citep{Schwartz1979} and scale  \citep{Day1981InfantObjects} invariance was found in infants at 3-4 months;  viewpoint invariance seems to be present in infants as young as 3 months old \citep{Kraebel2006Three-month-oldProcedure, Bornstein1986FineInfants}.

How these invariances are acquired, however, is not clear. Our work raises two points.  Firstly, the proliferation of architectural modifications to obtain online invariances in CNNs (see Section \ref{relatedwork}), even though developed without any psychological considerations, suggests that ``innate'' mechanisms for invariances are needed in this context, which further suggests that a form of innate architecture is needed in  human vision. In fact, we show that special purpose architectures are unnecessary, and learning invariances to object transformation can be achieved by a generic mechanism that performs feature learning on 2D samples.  This may also apply to humans.

Secondly, our results with the same/different network raise the possibility that the learning mechanism responsible for acquiring these invariances could be based on the evaluation of the ``sameness" of objects following changes in their retinal projections due to transformation. This is consistent with the analogical reasoning research literature that suggests the ability to form abstract relationship amongst objects is linked to the ability to solve the same/different tasks \citep{Gentner2021, Premack1983}. This skill is acquired during the first months, fully developed in 7-months old infants, and strongly linked to higher form of abstractions  \citep{Hespos2021}. Importantly, this approach does not require the complex process of object categorization through supervised learning that clearly does not apply at these early ages.

Of course, the same/different task still requires the information of whether the two objects are in fact the same or not, and this might be provided by a variety of mechanisms that exploit the temporal continuity of a sequence of retinal projections from objects in the world following spontaneous movements (e.g. eyes or head movement) and, later in the developmental stage, the active manipulation of objects while viewing them. 
It is important to acknowledge that our training regime sidesteps some important issues, notably, the specific mechanisms for telling the network that two images are the same or different.  Nevertheless, our findings show that the same/different signal is sufficient for CNNs to support online invariance to transformations, and this should motivate more research into how the infant brain might extract this signal in order to learn online invariances to various transformations.

 Using CNNs as a models for the human brain is an ambitious project. Thus far, the claim that CNNs provide a promising model of human vision has largely been supported by statistical measures of similarity between unit activations in models and neural populations in visual cortex \citep{Zhuang2020}. We believe that, for CNNs to be useful models of the visual stream, they also need to posses a set of fundamental functional properties that characterize human vision. Our high level contribution is to show that one of these properties, invariance to object transformations, that is not present in the system by default, can be acquired, even in a self-supervised manner.



\section{Conclusion} \label{Conclusion}
Overall, the results presented in this work suggest that it is possible to acquire a wide set of online invariances by pretraining on an appropriate dataset, with as few as 50 objects. Invariance was acquired for scale, rotation, translation, brightness, contrast, and, to lower amount, viewpoint. Critically, online invariance was obtained both with supervised and self-supervised training. Even though these invariances observed in the latent space will not always manifest themselves in classification performance at present due to catastrophic interference, recent progress in addressing catastrophic interference in other contexts  may eventually overcome these difficulties. From a psychological viewpoint, our findings suggest that online invariances may by the product of learning rather than an innate endowment. From a machine learning perspectives, our findings highlight how recent architectural modifications of CNNs to support these invariances may be unnecessary.  Rather, the focus should be on training models on the relevant datasets.  


\section*{Acknowledgement}
This project has received funding from the European Research Council (ERC) under the European Union’s Horizon 2020 research and innovation programme (grant agreement No 741134).


\newpage
\newpage
\appendix
\section{5AFT for ETH-80}
\begin{figure}[!ht]
\centering
  \includegraphics[width=1\linewidth]{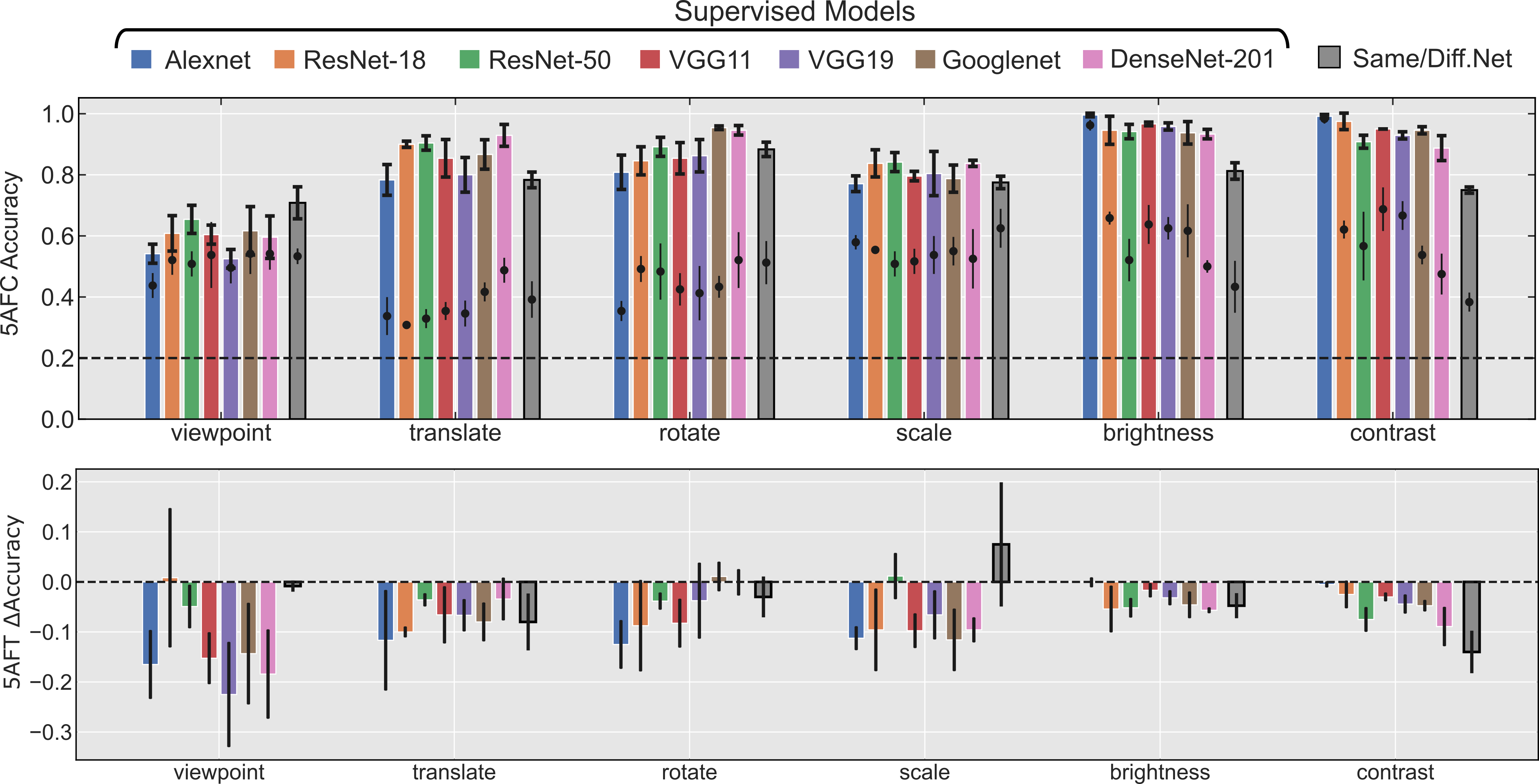}
\caption{\textbf{Top}: Results for the 5AFT task applied on the ETH-80, a novel dataset of photographed objects. The results are slightly worse than the tests on novel ShapeNet classes (Figure \ref{fig:FigBarplot}), which is expected given the different appearance of the samples, but the models are still showing a high degree of invariance. \textbf{Bottom}: We plot the differential accuracy between the 5AFT on the ETH-80 dataset and on the ShapeNet dataset. The difference is computed across pairs of seeds.}  
  \label{fig:APPFigBarplot}
  \end{figure}

We present here the result of the 5-alternative forced choice task (5AFC) run on the ETH-80 dataset. The details of the test were identical for the ShapeNet test. The results, Figure \ref{fig:APPFigBarplot}, top show an overall drop in accuracy averaged across networks of $6.5\pm7.8\%$. The variability in generalization across datasets varied between the different transformations and networks, but we could not find any pattern. Notice that even with the sharpest drop (with viewpoint transformation), the accuracy remained far above chance (which with 5AFT is 20\%). We directly show the difference in 5AFC accuracy between the ETH-80 and the ShapeNet results in Figure \ref{fig:APPFigBarplot}, bottom. The most striking results may be the drop in accuracy in the contrast transformation for the Same/Different Network. At the same time, the Same/Different network seems to generalize consistently well (across the 3 seeds) on new viewpoints.

\section{Invariant Representations Analysis} \label{SUPPinvanalysis}
\begin{figure}[!ht]
\centering
  \includegraphics[width=1\linewidth]{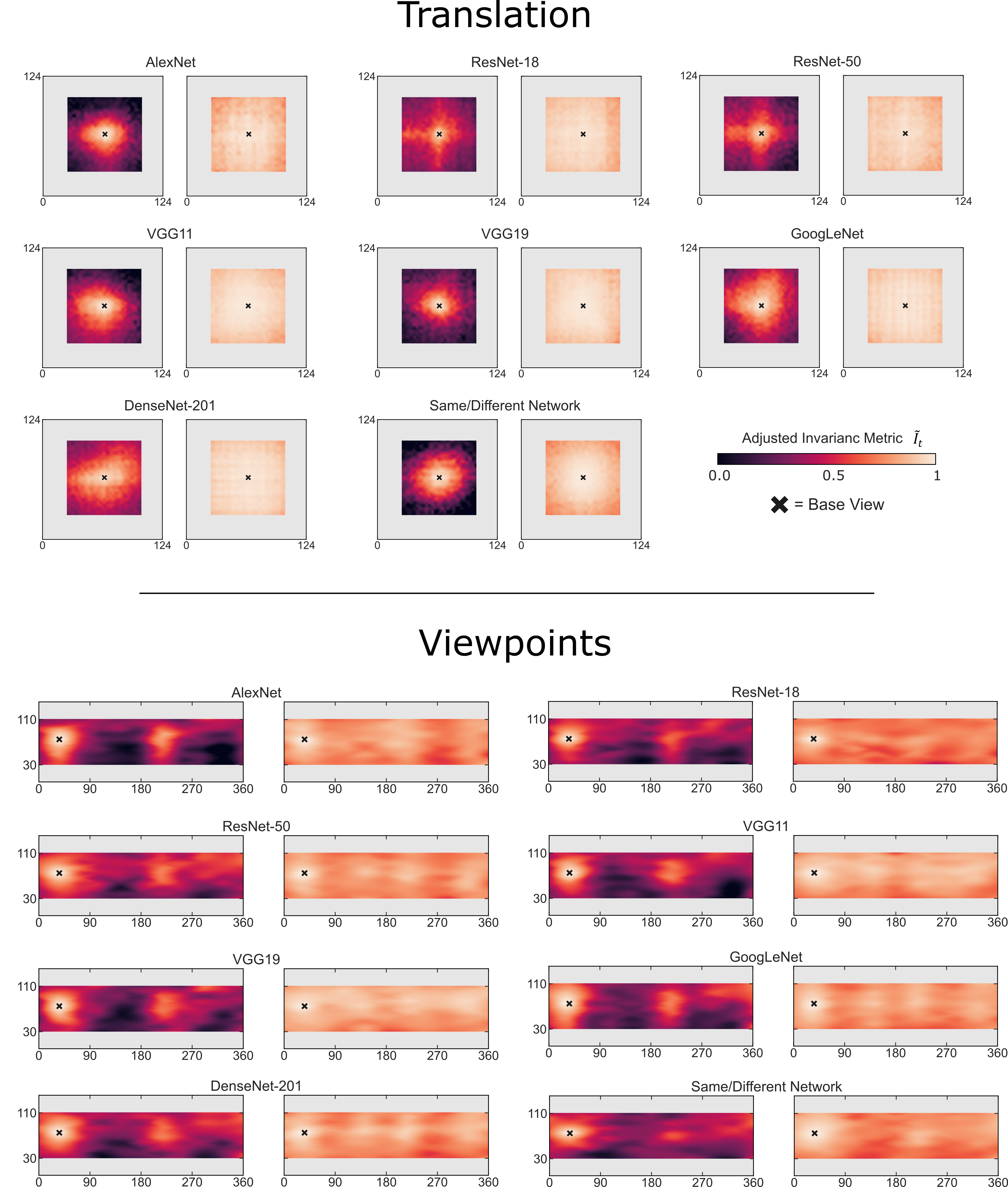}
\caption{The value of the Adjusted Invariance Metric for for the Translation and Viewpoint transformation conditions, for all networks. Transformation of objects across novel classes are compared to a base view (the \textbf{$\times$} in the plot), and the invariance representation across those poses is computed. For each network, we show \textbf{on the left} the result for the network trained on un-transformed objects, and \textbf{on the right} results for the network trained on transformed objects (either translation or viewpoints). }  
  \label{fig:APPallTranslVP}
  \end{figure}

We provide here the full results of the invariant representation analysis with the Adjusted Invariance Metric $\tilde{I}_t(\theta)$ for the translation and viewpoint transformations (which were presented only for two networks in the main text in Figure \ref{fig:FigInvariance}) in Figure \ref{fig:APPallTranslVP}. The results are averaged across 3 seeds. There is very little difference across networks, which similar patterns across un-transformed and transformed networks. The translation invariance results are consistent with \cite{Biscione2021JMLR}, which tested for this transformation across different datasets.

\subsection{Analysis of $I_t(\theta)$ and $U_t(\theta)$}
\begin{figure}[!ht]
\centering
  \includegraphics[width=1\linewidth]{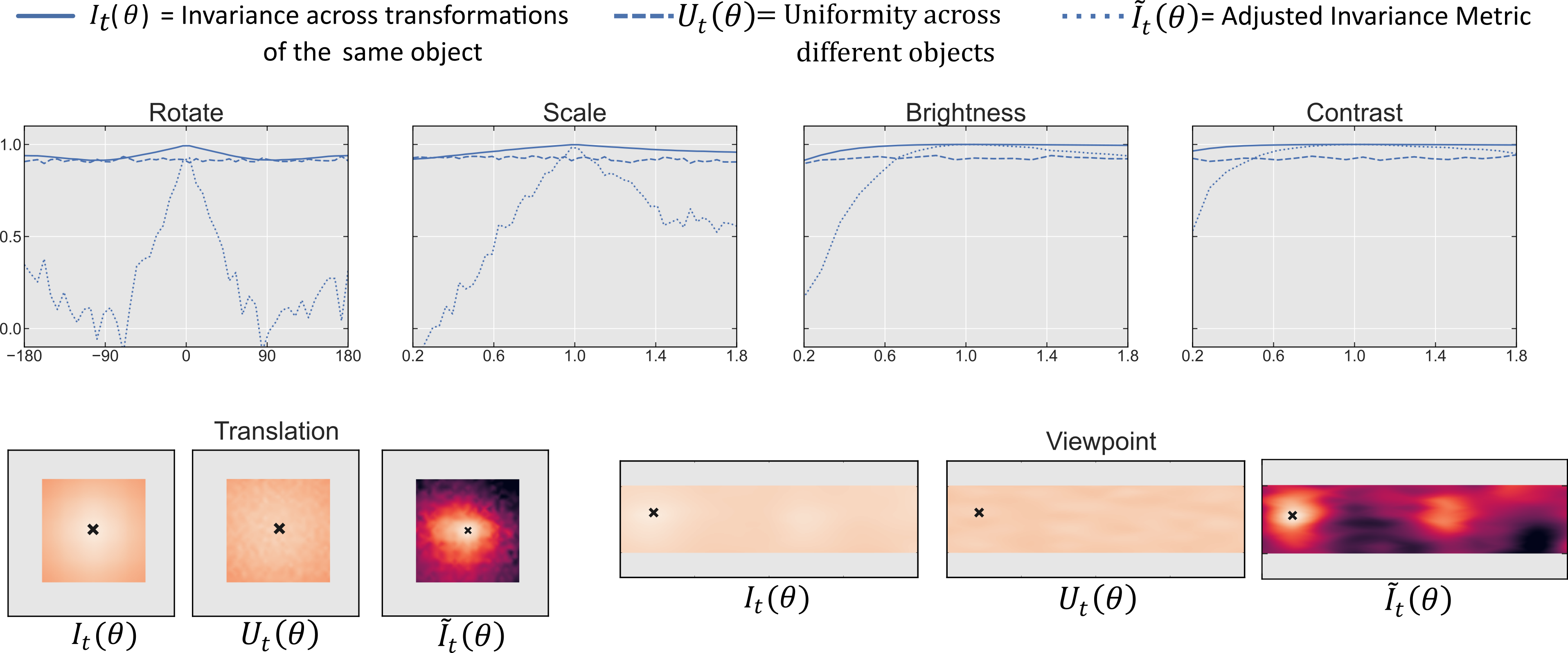}
\caption{Invariance $I_t(\theta)$, Uniformity $U_t(\theta)$ and Adjusted Invariance Metric $\tilde{I}_t(\theta)$ for a vanilla (untrained) VGG11. This is a clear case where the network shows invariance to transformation by trivially collapsing \emph{all} representations together, regardless of them belonging to the same or different objects. The metric $\tilde{I}_t(\theta)$ correctly adjusts the values to show that the network does not possess human-like invariance.}  
  \label{fig:SUPPL_vanilla}
  \end{figure}

In the main text, we defined $\tilde{I}_t(\theta)$ as the Adjusted Invariance Metric, a metric that uses the cosine similarity at the last layer to measure the invariance to object transformations and also accounts for the similarity across objects. To do that, we computed $I_t(\theta)$, the representation similarity across objects' transformations, and $U_t(\theta)$, the similarity across different objects. These two metrics are interesting on their own, as they reflect two different properties that the networks can acquire: $I_t(\theta)$ is \emph{Invariance} of representations under transformations, $U_t(\theta)$ can be seen as \emph{Uniformity} of representations across objects. Intuitively, a good model of the human visual system should possess a high Invariance within transformations and a low Uniformity between (different) objects. A striking example where both metrics need to be considered is a vanilla (untrained) network, which possess a high Invariance but also a high Uniformity (Figure \ref{fig:SUPPL_vanilla} for VGG11). A vanilla network tends to collapse all representations to the same output (thus returning the same class for any input) regardless of objects' transformation, identity, or category. It is, in some way, perfectly invariant to transformation, but it achieves that by trivially being invariant to everything, and it is thus a poor model of human object representation. Figure \ref{fig:SUPPL_vanilla} also shows how the Adjusted Invariance Metric $\tilde{I}_t(\theta)$ accounts for this.

\begin{figure}[!ht]
\centering
  \includegraphics[width=0.95\linewidth]{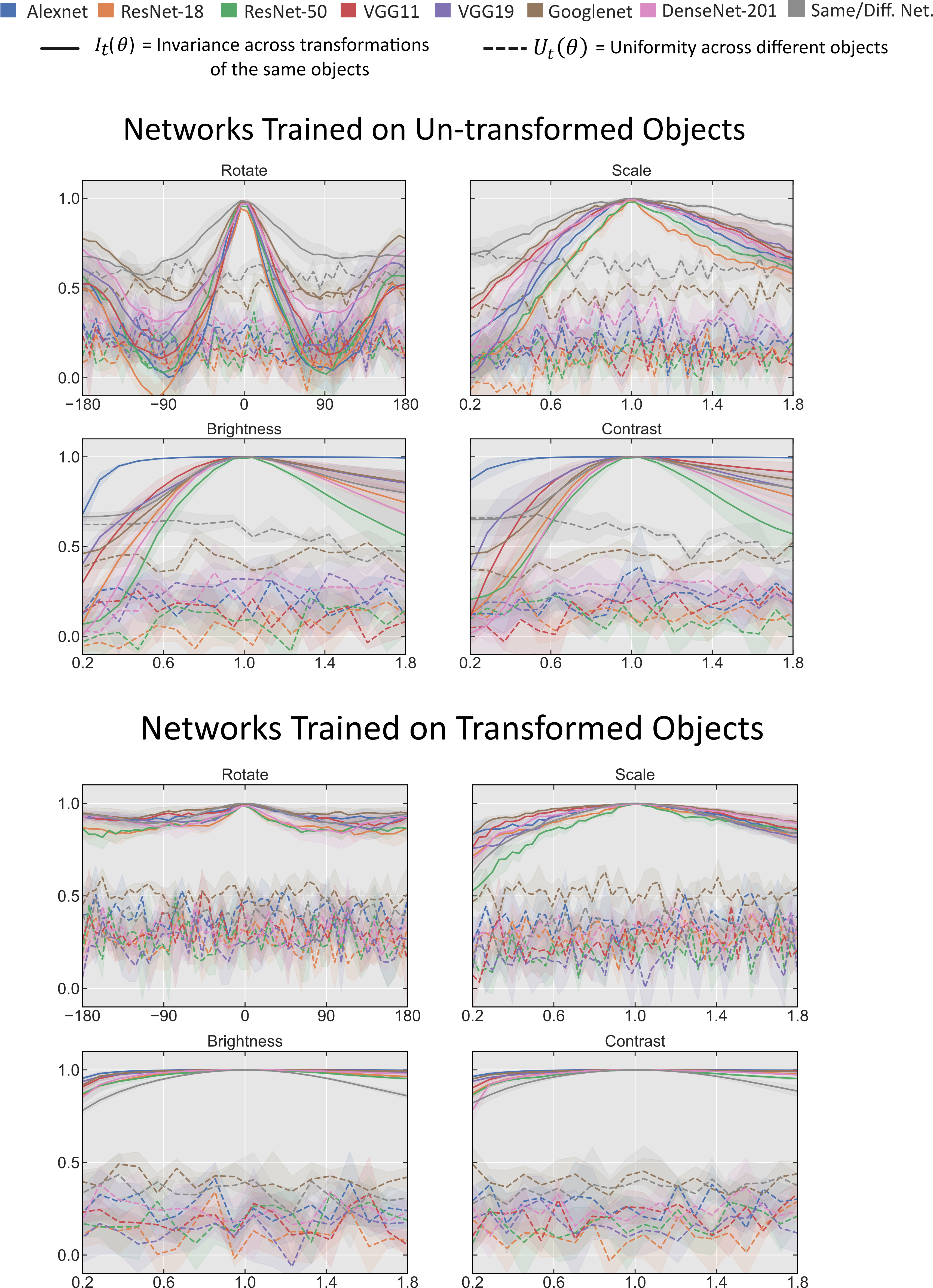}
\caption{Invariance ($I_t(\theta)$, continuous lines) and Uniformity ($U_t(\theta)$, dashed lines) for all networks, for rotation, scale, brightness and contrast transformations}  
  \label{fig:SUPPL_rsbc}
  \end{figure}

\begin{figure}[!ht]
\centering
  \includegraphics[width=0.9\linewidth]{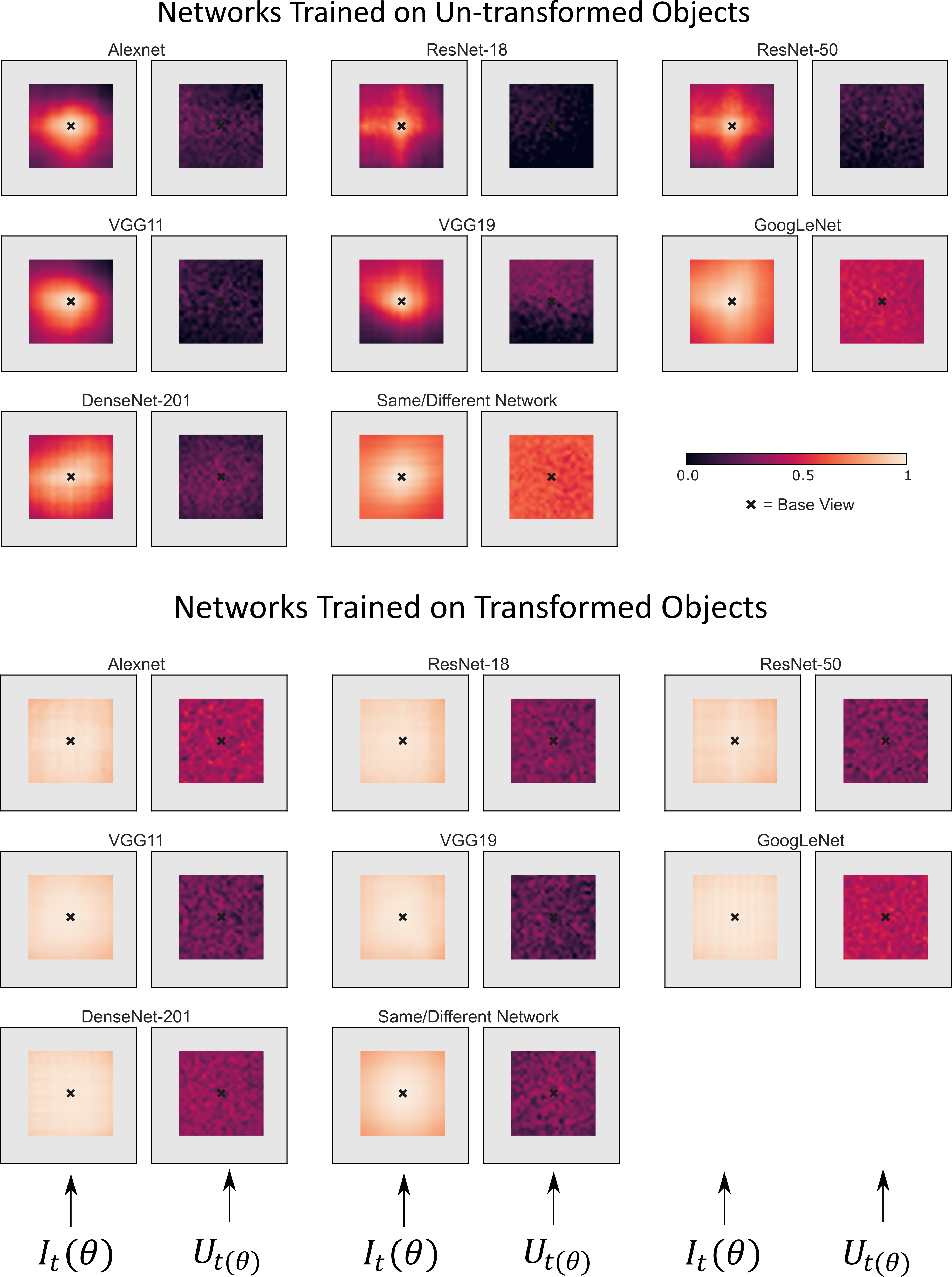}
\caption{Invariance ($I_t(\theta)$, on the left of each pair of heatmaps) and Uniformity ($U_t(\theta)$, on the right of each pair of heatmaps) for translation.}  
  \label{fig:SUPPL_translID}
  \end{figure}

\begin{figure}[!ht]
\centering
  \includegraphics[width=0.9\linewidth]{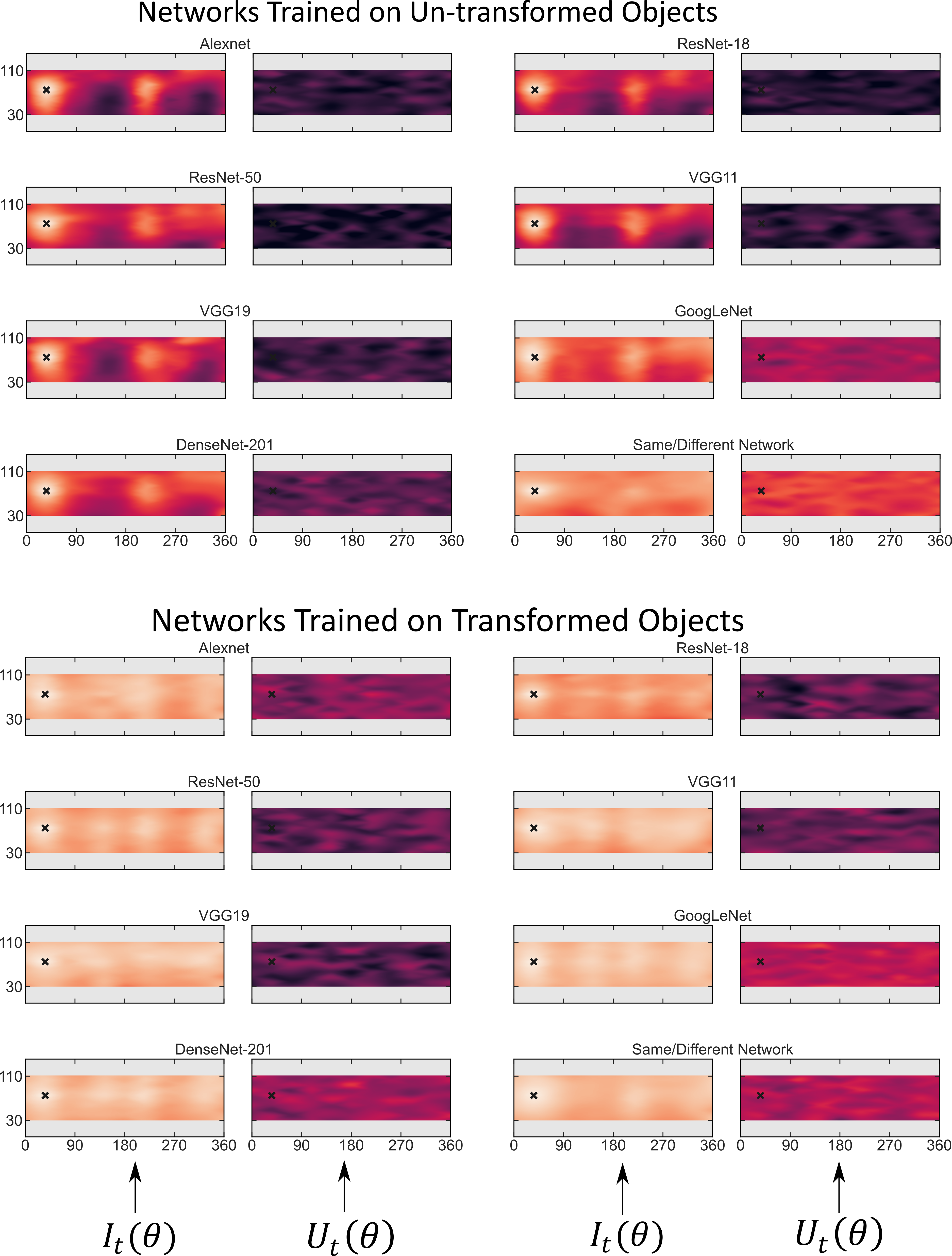}
\caption{Invariance ($I_t(\theta)$, \textbf{on the left} of each pair of heatmaps) and Uniformity ($U_t(\theta)$, \textbf{on the right} of each pair of heatmaps) for change in viewpoint.
Overall, networks trained on transformed objects (\textbf{bottom}) showed high Invariance and low Uniformity. Network trained on un-transformed objects (\textbf{top}) showed for most cases low Invariance and low Uniformity. In some cases, however, they showed fairly high Invariance, but also high Uniformity, which would produce unnatural internal representation in which different objects would be represented through the same activation pattern.}
  \label{fig:SUPPL_VPID}
  \end{figure}

We show these two metrics separately for the networks used in the main text in Figures \ref{fig:SUPPL_rsbc}, \ref{fig:SUPPL_translID}, and \ref{fig:SUPPL_VPID}. 
The network trained on transformed objects appears to have a high Invariance combined with a low Uniformity (Figures \ref{fig:SUPPL_rsbc}, \ref{fig:SUPPL_translID}, and \ref{fig:SUPPL_VPID}, bottom): transformed versions of the same objects are represented similarly, and different objects are represented separately. The pattern for networks trained on un-transformed objects was more peculiar (Figures \ref{fig:SUPPL_rsbc}, \ref{fig:SUPPL_translID}, and \ref{fig:SUPPL_VPID}, top). In most cases, the Invariance and Uniformity were both low, meaning that the network incorrectly represented different transformation of the same objects as different objects, but also that the network correctly separated different objects. However, in few cases, the Invariance was high (e.g. the translation and Viewpoint Invariance for the Same/Different Network or for GoogLeNet in Figure \ref{fig:SUPPL_VPID}, top), but the Uniformity was also fairly high, indicating that the network collapsed together representations of \emph{different}, transformed objects. For these networks, \emph{different} objects seen from different viewpoints have a \emph{similar} internal representation. Therefore, by analysing $I_t(\theta)$ and $U_t(\theta)$ separately, we can investigate different ways in which a network could fail to possess human-like invariance to transformations (that is invariance to transformation that does not \emph{also} collapses all representations together).

\section{Discrepancies with previous experiments}
In a somewhat similar experiment, \cite{Hernandez-Garcia2019LearningCategorization} found a low similarity across transformed samples, which is in direct contradiction with our findings. This can be accounted by the usage of different metrics: \cite{Hernandez-Garcia2019LearningCategorization} used Euclidean distance, whereas we used cosine similarity. 

\begin{figure}[!ht]
\centering
  \includegraphics[width=1\linewidth]{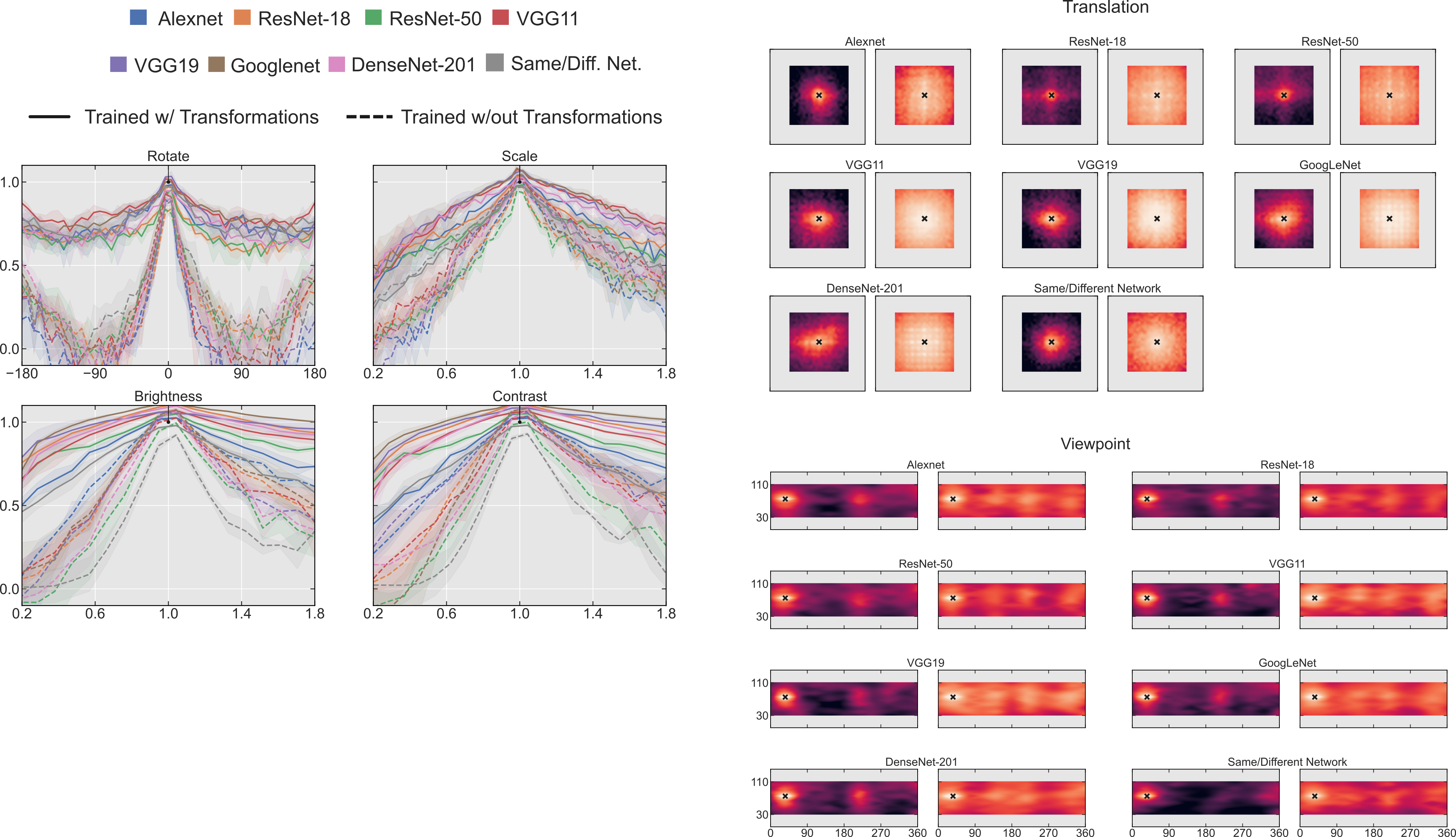}
\caption{Adjusted Invariance Metric $\tilde{I}_t(\theta)$ with Euclidean Distance in place of cosine similarity for all transformations and all networks. Euclidean distance is sensitive to absolute values of layers' activations, which results in underestimating the amount of invariance across transformations. For the panels on the right side, the pair of heatmaps for each network represents the invariance for network trained on un-transformed objects (\textbf{on the left}) and network trained on transformed objects (\textbf{on the right}).}  
  \label{fig:SUPPL_euclidean}
  \end{figure}

When using the Euclidean distance, change in the magnitude of the feature maps' activations (e.g. when using transformations such as brightness or contrast) will result in a low similarity score. The cosine similarity is, instead, normalized by magnitude, and is therefore not affected by an overall scaling of the activation values. We show that, similarly to \cite{Hernandez-Garcia2019LearningCategorization}, we also obtained lower invariance when using the Euclidean distance in computing the Adjusted Invariance Metric $\tilde{I}_t(\theta)$ in place of cosine similarity (Figure \ref{fig:SUPPL_euclidean}, compare this figure with Figure \ref{fig:FigInvariance}). Note that there is still a fairly strong separation between similarity in networks trained on transformed objects and networks trained on un-transformed objects, which was not measured in \cite{Hernandez-Garcia2019LearningCategorization}. 
The results obtained in the present work show that cosine similarity is a better suited metric for capturing invariance to transformations.  

\cite{Xu2021} similarly used the Euclidean distance to compute the dissimilarity across categories (building a Representation Dissimliarity Matrix, RDM) at two levels of a translation transformations (with objects either at the top or at the bottom of a canvas). They then computed the Representation Similarity Analysis (RSA) across the RDMs for a wide variety of CNNs and found low correlation across transformation. Their analysis is, therefore, very different to our approach, as they were focused on \emph{consistency} of representations across transformations, not \emph{invariance} of representations. Their finding of low consistency is nevertheless puzzling, and could be explained by several differences between ours experiments: firstly, like in \cite{Hernandez-Garcia2019LearningCategorization}, they relied on the Euclidean distance which has the problem discussed above. Moreover, they used networks pretrained on ImageNet. \cite{Biscione2020} showed that ImageNet, when trained with specific augmentation (that is, "Random Crop"), could provide a certain degree of translation invariance, but it is not clear whether this augmentation was actually used and whether it would be enough to generalize to such different samples. We hypothesise that pretraining the networks on a fully translate dataset, and using the cosine similarity instead of Euclidean distance, could possibly result in a high RSA, showing not only invariance, but consistency of representation across transformations.




\clearpage
\bibliographystyle{elsarticle-harv} 
  \bibliography{references}
\end{document}